\documentclass[11pt,a4paper]{article}
\pdfoutput=1
\usepackage{times,latexsym}
\usepackage{url}
\usepackage[T1]{fontenc}
\usepackage[nohyperref,acceptedWithA]{tacl2021v1}
\usepackage{etoolbox}
\usepackage[hidelinks=true]{hyperref}
\usepackage{appendix}
\usepackage{natbib}
\usepackage[utf8x]{inputenc}
\usepackage{booktabs}
\usepackage[disable]{todonotes}
\usepackage{adjustbox}
\usepackage{microtype}
\usepackage{multirow}
\usepackage{placeins}
\usepackage{longtable}
\usepackage{enumitem}
\usepackage{amsmath}
\usepackage{cleveref}
\usepackage{wrapfig}
\usepackage{breqn}
\usepackage{amssymb}
\usepackage{balance}

\begin{document}

\title{Learning English with Peppa Pig}

\author{Mitja Nikolaus\\
  Aix-Marseille University\\
  \texttt{mitja.nikolaus@univ-amu.fr}
  \And
  Afra Alishahi\\
  Tilburg University\\
  \texttt{a.alishahi@uvt.nl}
  \AND
  Grzegorz Chrupała\\
  Tilburg University\\
  \texttt{grzegorz@chrupala.me}}

\date{}

\maketitle
\begin{abstract}
  Recent computational models of the acquisition of spoken language
  via grounding in perception exploit associations between the spoken
  and visual modalities and learn to represent speech and visual data
  in a joint vector space. A major unresolved issue from the point of
  ecological validity is the training data, typically consisting of
  images or videos paired with spoken descriptions of what is
  depicted. Such a setup guarantees an unrealistically strong
  correlation between speech and the visual data.  In the real world
  the coupling between the linguistic and the visual modality is
  loose, and often confounded by correlations with non-semantic
  aspects of the speech signal. Here we address this shortcoming by
  using a dataset based on the children's cartoon {\it Peppa Pig}.  We
  train a simple bi-modal architecture on the portion of the data
  consisting of dialog between characters, and evaluate on segments
  containing descriptive narrations. Despite the weak and confounded
  signal in this training data our model succeeds at learning aspects
  of the visual semantics of spoken language.
\end{abstract}

\section{Introduction}
\label{sec:intro}

Attempts to model or simulate the acquisition of spoken language via
grounding in the visual modality date to the beginning of this century
\citep{roypentland2002learning} but have gained momentum recently
with the revival of neural networks
\citep[e.g.][]{synnaeve2014learning,harwath2015deep,
  harwath2016unsupervised,chrupala-etal-2017-representations,alishahi-etal-2017-encoding,harwath2018jointly,Merkx2019,havard2019models,rouditchenko2020avlnet,khorrami_2021,peng2021fastslow}.
Current approaches work well enough from an applied point of view, 
but most are not generalizable to real-life situations that humans or 
adaptive artificial agents experience. Commonly used training data
consist of images or videos paired with spoken descriptions
of the scene depicted: however, the type of input that a language learner receives 
from the environment is much more challenging.  
Firstly, speech is only loosely coupled with the visual modality in naturalistic settings
 \citep{matusevych2013automatic, beekhuizen2013word}. Speakers often mention 
 concepts that are not present in the immediate perceptual context, 
 or talk about events that are remote in space and/or time (for
 example past experiences or future plans).
 
Secondly, in addition to
correlations between the visual scenes and the {\it meaning} of spoken
utterances, there are also correlations with non-semantic aspects of
the speech signal, such as the voices of specific speakers, as well
as with non-speech ambient sounds. Although it is plausible that such
non-semantic correlations can sometimes be useful to the learner in
the general endeavor of making sense of the world, for the specific
task of learning the semantics of linguistic units they are likely more
often an obstacle, as they make it harder to zoom in on the
meaning-bearing aspects of the audio signal.

In the current study we make a first step towards simulating the
acquisition of language via grounding in perception in a more
naturalistic scenario.  Our main focus is on learning the meaning of
linguistic expressions from spoken utterances grounded in video.  We
use the well-known children's cartoon {\it Peppa Pig} as a case
study. Compared to commonly used video datasets, this dataset has a
number of interesting characteristics.  The visual modality is very
schematic, the language is simple in terms of vocabulary size and
syntactic complexity, and analysis of its linguistic features suggests
its suitability for beginner learners of English
\cite{kokla2021peppa,scheffler2021peppa}.  Crucially, however, most of
the speech in the videos consists of naturalistic dialogs between the
characters in which they do not only discuss the here and now, but
also often use displaced language.\footnote{For example, when Daddy
  Pig explains that they need to clean up before Mummy Pig sees the mess
  Peppa and George made, or when talking about plans to visit
  friends.}  Thus, the utterances are only loosely and noisily
correlated to the scenes and actions depicted in the videos.

This choice of data thus allows us to directly address the ecological limitations 
of the current approaches. In addition, the cartoon videos also contain 
comments interjected by the narrator. We use these for evaluating the 
acquisition of meaning as they are more descriptive and less noisy and allow 
us to measure performance, while controlling for speaker characteristics.

We implement a simple bi-modal architecture which learns spoken
language embeddings from videos, and train it on the Peppa Pig dataset.
Our contributions are the following:
\begin{itemize}
\item We evaluate model performance in terms of video fragment
  retrieval and additionally design controlled evaluation
  protocols inspired by the intermodal preferential looking
  paradigm \citep{hirsh1996intermodal};
\item We carry out ablations of model components in order to
  understand the effects of pre-training for the audio and video
  encoders, the role of temporal information, and of segmentation
  strategies while training. 
\end{itemize}
We show that despite the challenges of our naturalistic training data,
our model succeeds at learning associations between the form of spoken 
utterances and their visual semantics. Moreover, even though the model 
rarely hears words in isolation, it captures aspects of the visual meaning 
of frequent nouns and verbs.
Our ablation studies suggest that temporal information contributes
to video modeling (especially for longer segments), and that self-supervised pre-training
followed by fine-tuning of the audio encoder is key to the best
performance.

\section{Related Work}
\label{sec:related}

Early attempts at simulating grounded language learning focus on
interactions between adults and young children while playing with a
set of objects from different categories \cite{roy1999learning,roy2002learning,
  gorniak2003visually,mukherjee2003visual}. In a representative study
from this series, \citet{roypentland2002learning} use speech recorded from
such interactions paired with different views of the visible objects
to identify linguistic units (i.e.\ words) and visual categories, and
to map these two modalities together. A hard-coded visual system
extracts object representations from images, and spoken utterances are
represented as phoneme probabilities generated by an RNN pre-trained on
spectrograms.  Their experiments on small-scale data (around 20 words
and seven visual categories) show that the model can segment words and
map them to visual categories.

\subsection{Spoken Language Grounded in Images}
\label{sec:images}
The availability of datasets of images associated with spoken captions
such as Flickr Audio Captions \cite{harwath2015deep}, Places
\cite{zhou2014learning} and Spoken COCO \cite{hsu2019transfer} led to
a rapid development of neural models of grounded language learning; see
\citet{chrupala-visually-2021} for a comprehensive overview. In contrast to 
earlier approaches, these models are trained end-to-end directly on
large datasets.

Following the architecture proposed in \citet{karpathy2014deep} the visual and 
speech modality are usually encoded using separate pathways, and subsequently 
mapped into a joint representation space.
Visual features are extracted from a pre-trained
image classification model that processes the whole or a specific
region of an image (however see \citet{harwath2018jointly}, who train the
model end-to-end on images and their spoken captions on the Places
dataset). The audio encoder component in most models is 
either an adaptation of \citet{harwath2016unsupervised} which feeds a
spectrogram of the speech signal to a convolutional architecture, or a
hybrid architecture of convolutional followed by recurrent layers using
Mel-Frequency Cepstral Coefficient (MFCC) features from the audio
signal as input, as introduced by \citet{chrupala-etal-2017-representations}.

The majority of models of speech grounded in images are optimized for and evaluated on
image retrieval via spoken caption and vice versa. Additionally, a range of
diagnostic analyses have been performed on the hidden
representations of these models to study whether they encode 
the identity and boundaries of subword units such as phonemes
and syllables \cite{alishahi-etal-2017-encoding, harwath2019towards,
  khorrami_2021} as well as individual words
\cite{chrupala-etal-2017-representations,havard2019word}. Moreover, in
addition to examining form-meaning associations at the utterance
level, \citet{harwath-glass-2017-learning} explicitly learn a lexicon by
extracting audio and image segments, clustering each modality
separately, and mapping them together by calculating the pairwise
similarities of their members in the joint semantic space.

\subsection{Spoken Language Grounded in Video}
\label{sec:video}
There have also been recent attempts to learn spoken language grounded
in video instead of static images.  \citet{boggust2019grounding}
sample audio-visual fragments from cooking videos;  their
grounded model treats video frames as still images ignoring the temporal dimension.
\citet{rouditchenko2020avlnet} integrate the temporal information when
encoding videos from the Howto100m dataset \cite{miech2019howto100m},
and perform better than previous work in language and video clip
retrieval.

Models trained on instructional video datasets often do not
generalize well to other domains. \citet{monfort2021spokenmoments}
highlight this limitation and show that training on their larger and
more diverse Spoken Moments in Time dataset leads to better
generalization.  But the point remains that these video datasets contain
descriptive speech, thus ensuring that there is a strong correlation
between the spoken language and their visual context, a characteristic
that is not representative of the experience of learning language in
 real world. We remedy this limitation by using a video dataset that does 
 not guarantee a direct description of the visual context. 

\subsection{Child Language Learning from Video}
There are many studies on young children learning language by watching
videos; see \citet{vanderplank2010deja} for a survey. The main takeaway
of these studies is that language learning is much more effective in a
social, conversational setting than by passively watching videos
\cite{kuhl2003foreign,anderson2005television,robb2009just},
but learning does happen in such
contexts. Importantly for our goal, techniques such as the intermodal
preferential looking paradigm have been developed to systematically test young 
language learners' knowledge of words, syntactic structure and semantic roles
\cite{hirsh1996intermodal,bergelson20126,noble2011comprehension}.
\citet{nikolaus-fourtassi-2021-evaluating}
employ this evaluation strategy to test semantic knowledge at word and
sentence level in their computational model of word learning from
images. We adapt this approach to evaluate how our grounded model
associates semantic information to spoken words and utterances from
video.

\subsection{Intra-linguistic Statistics}
One further aspect of learning spoken language via visual grounding is
the fact that grounding is only part of the story. Human children
arguably infer substantial amounts of information about language
structure and meaning from purely intra-linguistic co-occurrence
statistics \citep[e.g.,][]{saffran1996statistical}. A similar mechanism is what 
allows 
written language models
such as BERT \citep{devlin-etal-2019-bert} or GPT-3 \citep{brown2020language} 
to capture and exhibit relatively sophisticated
linguistic knowledge. Loosely similar approaches have started to also
make an impact for the spoken modality
\citep[e.g.][]{wav2vec2,hsu2021hubert}. Here we take a simple
pre-training-based approach to integrating this type of
self-supervision with learning-via-grounding.

\section{Method}
\label{sec:method}

The main focus of this study is on the data and evaluation. We thus
keep the components of our architecture simple, and follow established
modeling practices whenever possible.

\subsection{Dataset}
We use the dataset provided by \citet{papasarantopoulos2021narration}
which consists of metadata for the set of 209 episodes (seasons 1--5) of the
English-language version of {\it Peppa Pig}.\footnote{We purchased the
corresponding Peppa Pig episodes on DVD support.}
The annotations created by \citet{papasarantopoulos2021narration}
feature written transcriptions aligned with the audio as well as
segmentation into {\it dialog} and {\it narration}.\footnote{The
  quality of the alignment and segmentation in this dataset is
  variable. In cases where exact alignment is needed, such as for
  word-level analyses, we re-align the transcriptions using
  \url{github.com/lowerquality/gentle}.}  Dialogs are the parts spoken
by the characters, while narrations are comments inserted by the
narrator, which are more descriptive in nature. All the narration
segments are uttered by the same voice actor. We use the dialogs for
training the model, and set aside the narrations for evaluation
purposes only. A small portion of the dialog data is also used for
validation.  Specifically, out of the total 209 episodes, we use
dialog from episodes 1--196 for training, and 197--209 for
validation. We set aside narrations from episodes 1--104 for
validation and 105--209 for testing. We disregard portions of the
video which are annotated as neither dialog nor narration: this means
our data consists mostly of video clips which contain some
speech.\footnote{Manual analysis of a random sample of 50 segments
  split according to the method described in \Cref{sec:preprocessing}
  showed that approximately 6\% of them contained no discernible words.}
\Cref{tab:ds-stat} shows the sizes of the training and validation
splits. The vocabulary size of transcriptions corresponding to the
training data is 5,580.

\begin{table}[htb]
  \centering \begin{tabular}{llrr}
	\toprule
	Split &      Type &  Size (h) &  \# Clips \\
	\midrule
	train &    dialog &     10.01 &    15666 \\
	val &    dialog &      0.66 &     1026 \\
	val & narration &      0.94 &     1467 \\
	test & narration &      0.64 &     1006 \\
	\bottomrule
\end{tabular}

  \caption{Duration in hours and number of clips (\textsc{fixed} condition) for 
  all dataset splits.}
  \label{tab:ds-stat}
\end{table}

\subsection{Preprocessing}
\label{sec:preprocessing}
Our model is trained to discriminate positive video-audio pairs from
negative ones.  The positive pairs are those that are temporally
coincident in the original video file. In order to generate these
training items we need to split the videos into fragments.  When
preparing training data, we use annotations to separate dialog and
narration data, but we \emph{do not} use alignment with
transcriptions for further segmentation, in order to make the setting
naturalistic. Processing long segments of video and audio is not
tractable on commodity GPU hardware, and we thus segment the data into
brief snippets roughly comparable in length to the duration of a short
sentence or a phrase. We use the following two segmentation
strategies:

\paragraph{Fixed} Using this approach we simply split sections into
fixed-length non-overlapping fragments of 2.3 second duration. This
length is close to the mean duration of audio aligned to a single line
of subtitles. The number of clips for each dataset split is shown in 
\Cref{tab:ds-stat}.

\paragraph{Jitter} In this approach the mean duration of the segments
is the same (2.3 seconds) but we randomly vary the length of the
video, and independently, of the corresponding audio around this
average duration. This means that (i) the segments can be partially
overlapping and (ii) the video and the audio it is paired with are
of different length. Specifically, we sample the fragment
duration $d$ (in seconds)
from the following distribution:
\begin{equation}
  d \sim \min(6, \max(0.05, \mathcal{N}(2.3, 0.5)))
  \label{eq:jitter}
\end{equation}
The video is subsampled to 10 frames per second, and to
$180\times 100$ resolution.\footnote{Performance is better with higher
  resolution, but it makes GPU memory
  requirements prohibitive.}  The audio is converted to mono by
averaging the two channels and the raw waveform is used as input. We
use the original sample rate of 44.1 kHz (instead of downsampling to
the 16 kHz sample rate used for pre-training \textsc{wav2vec2}) as we
found out that this helps with generalization performance on the
narration validation data.

For evaluation we have a number of different conditions and evaluation
metrics described in detail in \Cref{sec:eval} and in some of these
conditions we use the subtitles to guide
segmentation.

\subsection{Model Architecture}
\label{sec:model}
We adapt the general modeling framework from studies on spoken
image-caption data
\citep{harwath2016unsupervised,chrupala-etal-2017-representations}:
our objective function is based on a triplet-like contrastive loss with margin which
encourages the matching audio and video clips to be projected nearby in
the embedding space, and mismatching audio and video clips to be far
away:
\begin{dmath}
  \ell = \sum_{av}\left[\sum_{a'} \max(0, S_{a'v} - S_{av} +
    \alpha) + \sum_{v'} \max(0, S_{av'} - S_{av} + \alpha) \right]
  \label{eq:triplet}
\end{dmath}
where $\alpha$ is a margin, $S_{av}$ is a similarity score between a
matching audio-video clip pair, and $S_{a'v}$ and $S_{av'}$ denote
similarity scores between mismatched pairs, i.e.\ negative examples
from the current batch. Our heuristic to generate positive and
negative examples is very simple: we consider an example
positive if the audio is temporally aligned with a video clip in our
data. Other pairs of audio-video clips are considered negative.

\subsubsection{Audio Encoder}
The audio encoder portion of the model consists of a {\sc small
  wav2vec2} model \citep{wav2vec2} pre-trained in a self-supervised
fashion, \emph{without} any supervised fine-tuning.\footnote{Available from
  \url{https://dl.fbaipublicfiles.com/fairseq/wav2vec/wav2vec_small.pt}.}
The {\sc wav2vec2} architecture learns audio embeddings by
self-supervised learning driven by a contrastive loss applied to 
quantized latent representations of masked frames, loosely inspired by
the BERT approach to language modeling \citep{devlin-etal-2019-bert}.

The output of this module is a temporal sequence of 28-dimensional vectors. We
pool this output across time using an attention mechanism with
dimension-wise weights \citep{Merkx2019}:
\begin{equation}
  \begin{aligned}
    \mathbf{A} = & \mathrm{softmax}_t\left(\mathrm{MLP}(\mathbf{X})\right)\\
    \mathbf{z} = & \sum_t \left( \mathbf{A}_{t} \odot \mathbf{X}_{t} \right),
  \end{aligned}
  \label{eq:att-pool}
\end{equation}
where $\mathbf{X}$ is the tensor with the encoder output vectors for
each time-step $t$: an MLP followed by a time-wise
softmax is used to compute an attention weight for each time step and for each
dimension.
The pooling is followed by a linear projection to 512 dimensions and $L_2$
normalization. For our experiments we also use versions of the encoder
where the \texttt{wav2vec2} weights are frozen, as well as a randomly initialized
rather than pre-trained version.

\subsubsection{Video Encoder}
As a video encoder we use the 18-layer ResNet (2+1)D architecture
\citep{tran2018closer}, pretrained on the action recognition dataset
Kinetics-400 \citep{DBLP:journals/corr/KayCSZHVVGBNSZ17}. The
pre-trained model is available via Pytorch.\footnote{See
  \url{https://pytorch.org/vision/stable/models.html\#resnet-2-1-d}.}  This
architecture implements 3D convolution by decomposing it into a 2D
spatial convolution followed by 1D temporal convolution.  The output
of this module is aggregated using the attention mechanism with the
same architecture as for the audio module, linearly projected to the
same dimensionality as the audio (512) and $L_2$ normalized.  For our
experiments we also use a version of the video encoder without
pre-training.

\paragraph{\textsc{Static} baseline}
As a baseline to investigate the contribution of temporal information to
video modeling we swap the video ResNet (2+1)D with the 2D ResNet
pre-trained on ImageNet, which embeds each video frame
separately. These frame embeddings are then attention-pooled as with
the standard video encoder. 

To further investigate the impact of temporal information while 
controlling for model architecture, we evaluate model performance in a 
condition where we randomly scramble the video frames within a clip at test 
time, thereby removing any useful temporal information.

\subsection{Evaluation}
\label{sec:eval}
The most common approach to evaluation for visually grounded models
trained on spoken image captions is caption-to-image retrieval (often
combined with image-to-caption retrieval); this technique
has been carried over from text-based image-caption modeling.
 With the standard spoken caption datasets this approach is unproblematic since
the content of the captions is not correlated with extra-linguistic
clues in the speech signal, such as speaker identity (since speakers
are randomly assigned to captions) or non-speech environmental
sounds. In such an artificial setting, a retrieval metric measures the ability of the
model to match spoken utterances to images based on their semantic
content. This is not the case for the {\it Peppa Pig} dataset: here we
can expect that when a video segment depicts a particular character
(e.g.\ George) then the audio in this segment is more likely to contain
utterances spoken by the voice actor playing George.  Moreover, some 
characters might have a tendency to talk about certain topics more often 
than others, and the model might pick up on these associations instead of 
paying attention to the actual meaning of the uttered words.
%
Due to these factors, in a naive retrieval setting, a model
could obtain a high score by mostly capturing these non-linguistic
correlations.

In order to control for these factors we leverage the
narrator speech in the videos. These utterances are always spoken by
the same actor, so speaker identity cannot be used as a clue for
matching video and audio. Furthermore, the narration segments are akin
to video captions in that they tend to describe what is happening in
the video and thus their semantic content is more strongly
correlated with the content of the video than in the case of the
dialog, which is also a desirable feature for the purposes of system
evaluation.

\subsubsection{Video Retrieval}
\label{sec:retrieval}
For the retrieval evaluation, as for training, we also use both
the \textsc{fixed} and the \textsc{jitter} segmentation strategies;
however, for most conditions, we only report retrieval for the
\textsc{fixed} evaluation data.

We encode each audio clip in a candidate set sampled from the
validation (or test) data using the speech encoder part of the model;
similarly we encode each video clip using the video encoder. We then
measure cosine similarity between the encodings of the audio clip and
all the video clips. If the video clip corresponding to the audio is
among the $n$ most similar video clips, we count that as a
success. The proportion of successes across all audio clips gives us
the retrieval metric known as recall@$n$. In Section~\ref{sec:results} 
we report recall@$N$ of the complete model on narration test data 
for values of $N$ between 1 and 10; for the rest
of the experiments in this paper we focus on $n=10$. 
We set the candidate set size to $100$, and thus
the random baseline for the recall@10 is $10$\%. In order to quantify
uncertainty in this evaluation due to the test data we repeat this
procedure 500 times with randomly sampled candidate sets and visualize
the score distribution.

\subsubsection{Triplets}
\label{sec:triplets}
The absolute value the recall@10 of this metric may be hard to
interpret as it depends on the size and content of the candidate set.
For this reason, we evaluate model performance using a simpler,
controlled scenario, inspired by intermodal preferential looking
paradigms in child language acquisition
\citep{hirsh1996intermodal}. The proposed metric can be seen as a
multimodal version of the ABX score proposed in \citet{schatz2016abx}.

\begin{figure}
	\centering
	\includegraphics[width=\columnwidth]{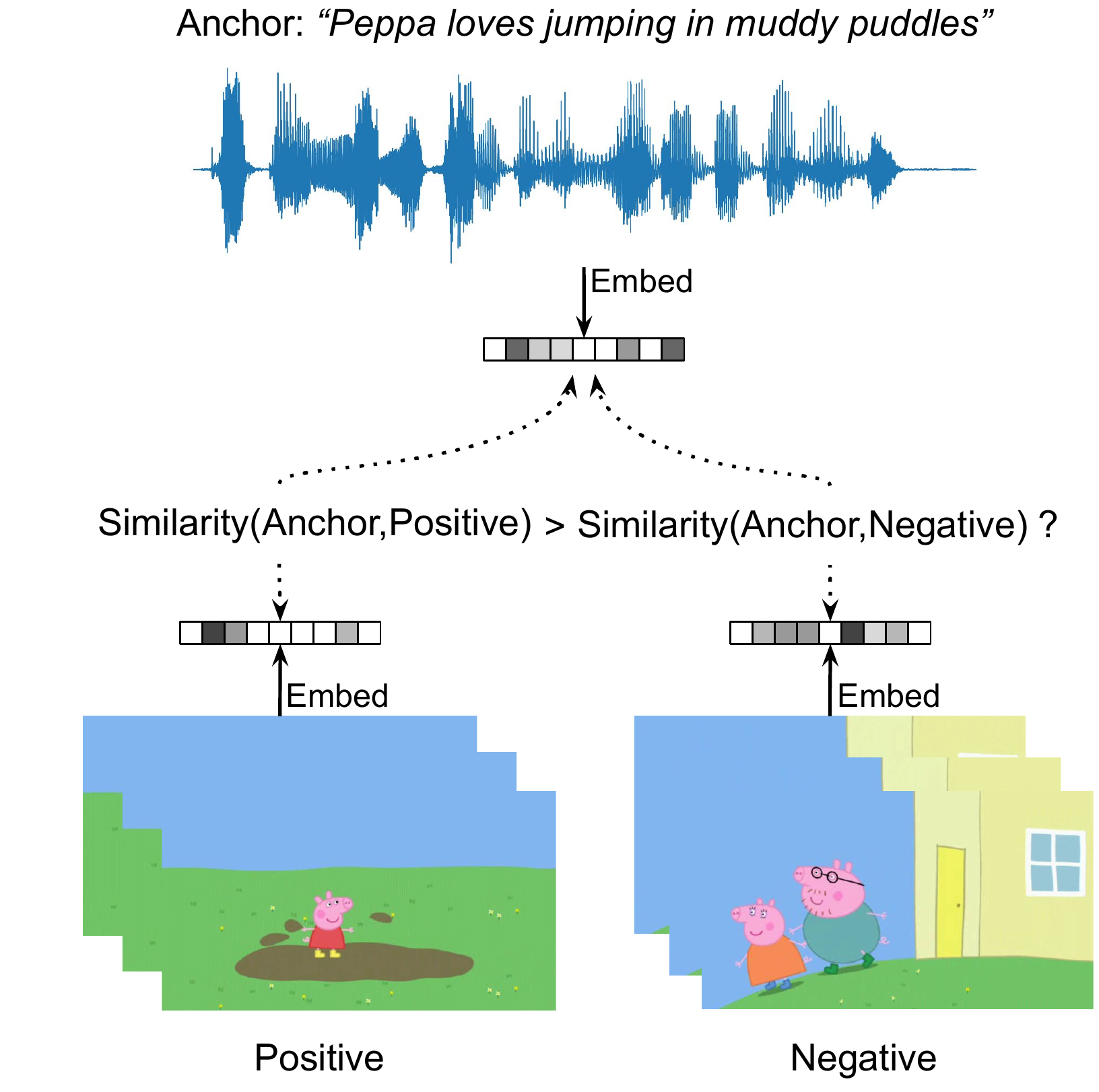}
	\caption{Triplets Evaluation: Given a reference audio sequence (anchor), we 
	measure the model's performance at choosing the matching video (positive) 
	over a random distractor video (negative).}
	\label{fig:triplets_eval}
\end{figure}

We extract clips aligned to a single subtitle
line, group them by length, and for each pair of same-length video
clips,\footnote{To keep test items independent, the pairing of video
  clips is done such that each clip only occurs as a member of a single
  triplet.} we extract the audio from one of them (selected at
random) -- this is our {\it anchor}. The video clip from which the
anchor was taken is the {\it positive} one, and the other video clip
is the {\it negative} one. This triplet of stimuli forms a single test
item.

We use the model's audio encoder to encode the anchor, and the
video encoder to encode both video clips. We then check whether 
the anchor is more similar to the positive or to the negative clip in terms of cosine
similarity (see \Cref{fig:triplets_eval} for an example).  More precisely, {\it triplet 
accuracy} is the mean over all triplets of the following quantity:
\begin{equation}
\hspace{-.03cm}
  \frac{\mathrm{signum}(\mathrm{cosine}(A, P) - \mathrm{cosine}(A, N)) + 1}{2}
  \label{eq:triplet-acc}
\end{equation}
where $A$ is the anchor, $P$ is the positive and $N$ is the negative video 
clip. 
For this metric, we expect random-guessing performance to be at 0.5, and perfect
performance to be at 1.0, regardless of the specific set of test items. We also 
quantify uncertainty by resampling the triplets $500$ times from the dataset, 
and display the score distribution.

\subsubsection{Minimal Pairs}
\label{sec:targeted}
While the triplet evaluation gives us a general idea about whether the model 
has learned a mapping between audio and video at the utterance level, it 
cannot tell us whether the model has acquired the grounded semantics 
of individual words.

To address this question, we probe the model's performance in a more targeted 
triplet setup, where the model is required to select the correct video from a 
pair of videos whose corresponding transcripts only differ in one target word.
To construct the evaluation set, we search the transcripts of the validation 
data for phrases with minimal differences with respect to the most commonly 
occurring nouns, verbs and adjectives. We set the minimum frequency of the 
target word in our training set to 10, and the minimum phrase duration to 0.3 
seconds.\footnote{For shorter sequences, we do not 
expect that the video contains enough semantic information to 
distinguish target and distractor. A phrase can also be a single word.}
Following \citet{nikolaus-fourtassi-2021-evaluating},
we pair every such triplet example with a corresponding counter-example to 
control the evaluation for linguistic biases in the dataset.

\begin{figure}[ht]
  \centering
  \includegraphics[width=\columnwidth]{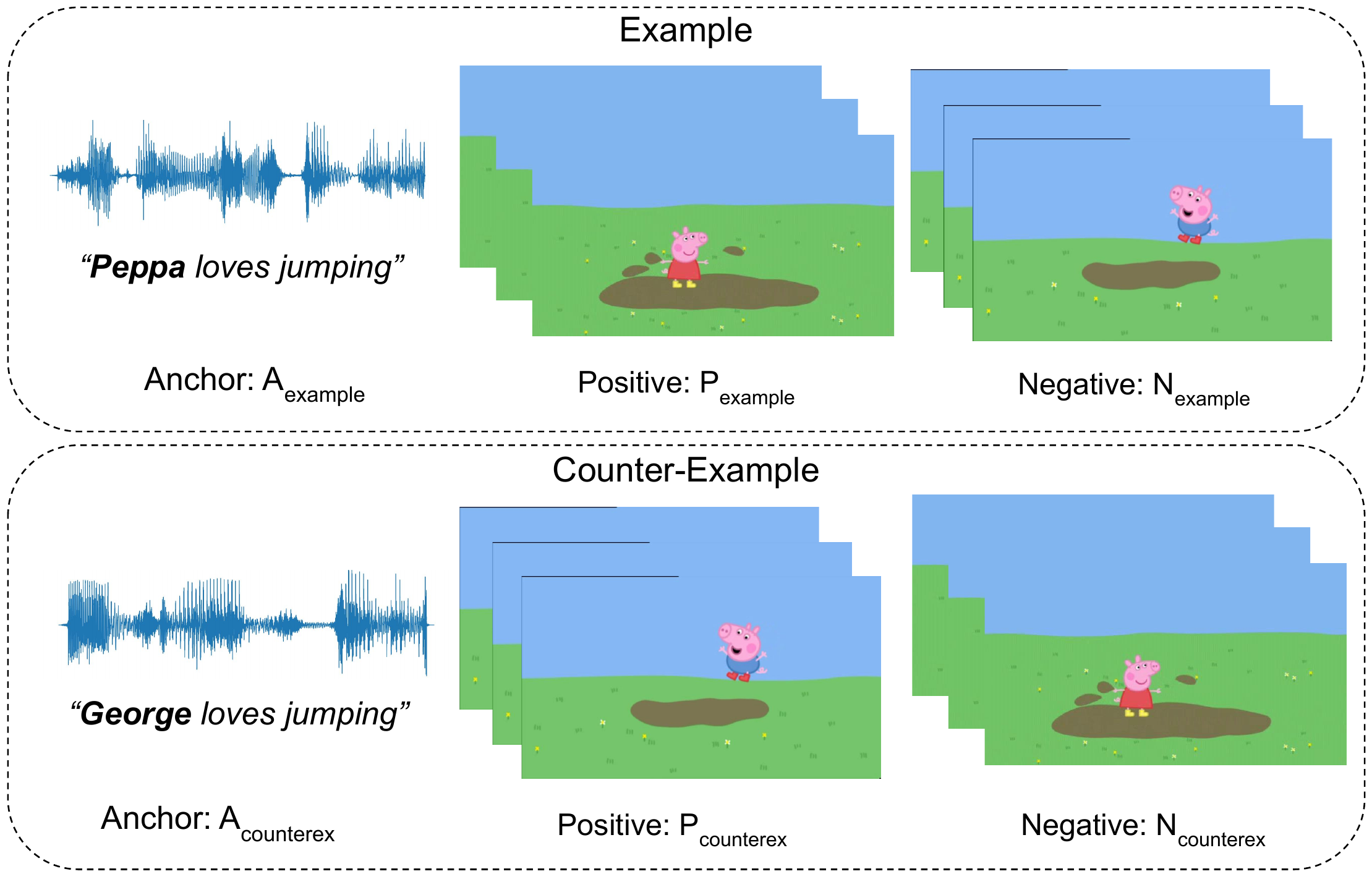}
  \caption{Example and counter-example triplets corresponding to minimal pairs {\it Peppa loves jumping} and {\it George loves jumping}.}
  \label{fig:minimal_pairs}
\end{figure}

\Cref{fig:minimal_pairs} shows an example of how two counter-balanced test
trials are constructed from audio and video clips. 
Here, the anchor $A_{\text{example}}$ of the example
triplet is the audio of \textit{Peppa loves jumping}, the positive video 
$P_{\text{example}}$ is the corresponding video, and the negative video 
$N_{\text{example}}$ is the video corresponding to \textit{George loves 
jumping}. In the counter-example triplet, the anchor $A_{\text{counterex}}$ is 
the audio of \textit{George loves jumping}, and the positive and negative videos 
are flipped.

We measure word accuracies by calculating the triplet accuracy 
for all triplets that contain a given word (e.g.\ 
\textit{Peppa} in the previous example) either as target or distractor. 
That is, we take into account all cases where the model needs to use 
the meaning of the given word for either choosing or rejecting a video. 
We report word accuracy for all nouns and verbs for which we find at least 
100 pairs of triplets in the validation set. We did not find enough examples for any 
adjectives, and thus did not include them in our evaluation.

\section{Experimental Settings}
We implement the architecture in PyTorch \citep{NEURIPS2019_9015}. We
use the Adam optimizer \citep{kingma2014adam} with the scheduling
described in \citep{devlin-etal-2019-bert}. We train every
configuration on a single GPU and stop training after 48 hours, with
batch-size 8 and accumulating gradients over 8 batches, in 16 bit
precision mode. For each model configuration we save model weights
after each epoch and report results for the checkpoint which gets the
best triplet accuracy on the narration validation data. 

Our code is publicly available at \url{https://github.com/gchrupala/peppa},
and can be consulted for further details of the experimental setup.

\subsection{Sources of variability}
\label{sec:variability}
We account for two sources of variance in the results. Firstly, for
each model configuration we ran four separate training runs in order
to account for the effect of random initialization. Secondly, we
estimate the variance due to validation/test sample by 
resampling validation and test items 500 times. In the case of the minimal 
pairs evaluation, we employ bootstrapping with 100 re-samples. In most
cases in \Cref{sec:results}, we pool variance from both sources and report
overall spread, except when specifically focusing on the contribution
of each source.

\section{Results}
\label{sec:results}

\begin{figure}[htb]
  \centering
  \includegraphics[width=\columnwidth]{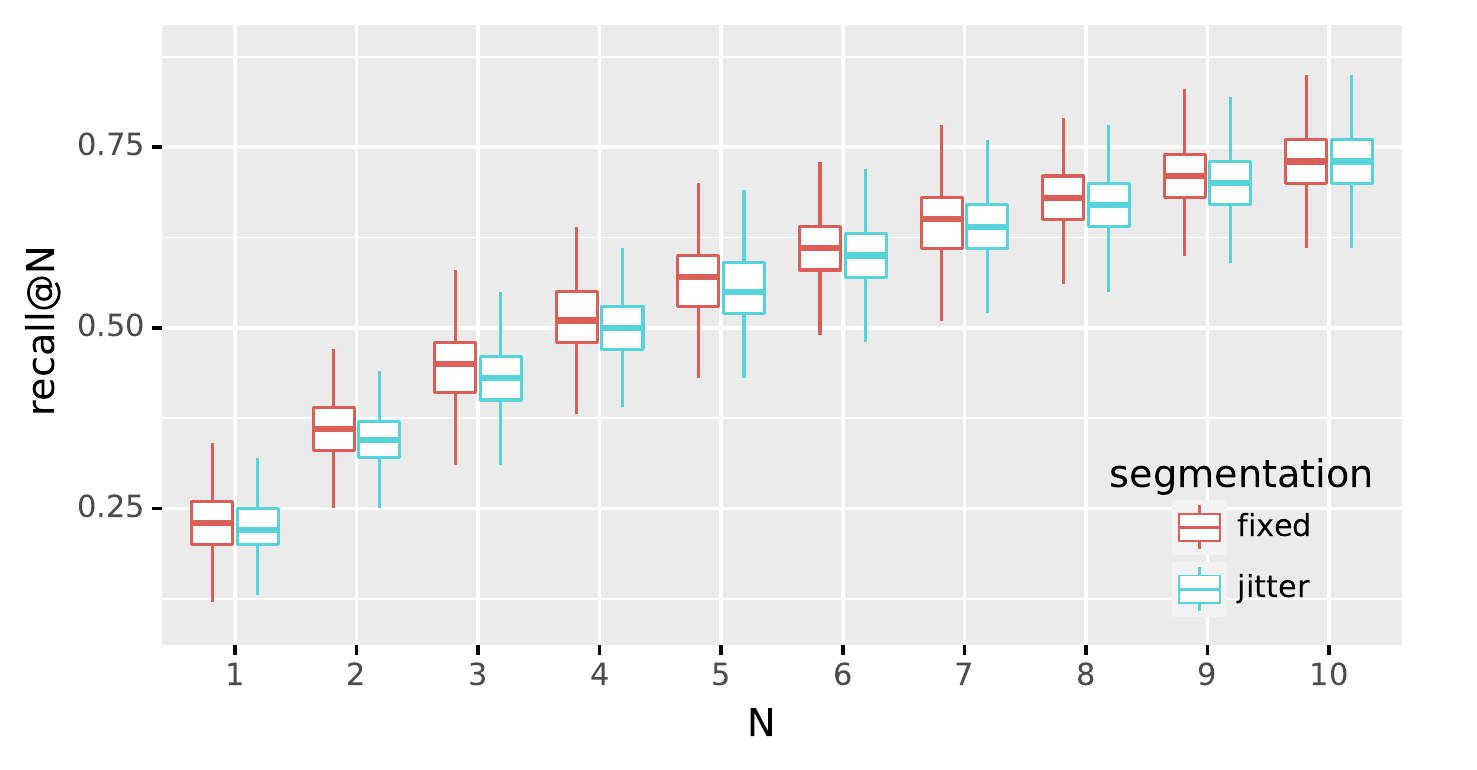}
  \caption{Recall@$N$ as a function of $N$, for the narration test
    data. We show recall for the complete model, for the {\sc fixed} and {\sc jitter}
    retrieval evaluation settings. }
  \label{fig:recall_at_1_to_n}
\end{figure}

\begin{table}[htb]
  \begin{tabular}{lll}
\toprule
R@10 (fixed) & R@10 (jitter) & Triplet Acc \\
\midrule
 0.73 ± 0.05 &   0.73 ± 0.04 & 0.91 ± 0.01 \\
\bottomrule
\end{tabular}

  \caption{Results of the full model on narration test
  	data. We show the mean and standard deviation of 
  	bootstrapped scores, pooled over four training runs
	(chance recall@10 = 10\%; chance triplet accuracy = 50\%).}
  \label{tab:test_scores}
\end{table}

\Cref{fig:recall_at_1_to_n} shows recall@$N$ for values of $N$
between 1 and 10 for the complete model on test narration data in both 
{\sc fixed} and {\sc jitter} conditions. Both plots show that the value of 
recall@$N$ increases monotnically. For the rest 
of this paper, we only report recall@10.

\Cref{tab:test_scores} presents the recall@10 and triplet accuracy
scores on test narration data obtained with the complete
model. In \Cref{sec:ablations} we investigate the impact
of various components of our training setup on performance as measured
by recall@10 and triplet accuracy.  In \Cref{sec:minimal-pairs} we
focus on the targeted evaluation via minimal pairs.

\subsection{Ablations}
\label{sec:ablations}
For completeness, we report results on both dialog and narration
data. However, the scores on narration are our main focus as they are
not confounded by speaker-based clues, and thus indicate to what
extent the model learns aspects of utterance meaning.

For experiments in Section~\ref{sec:pretraining} we include each run as a separate boxplot
to show the consistency of the results between runs in
different training conditions.  For the other experiments we collapse
the results of the four runs to avoid clutter.

\subsubsection{Pretraining and Fine-tuning}
\label{sec:pretraining}
\begin{figure*}[htb]
	\centering
	\includegraphics[width=\textwidth]{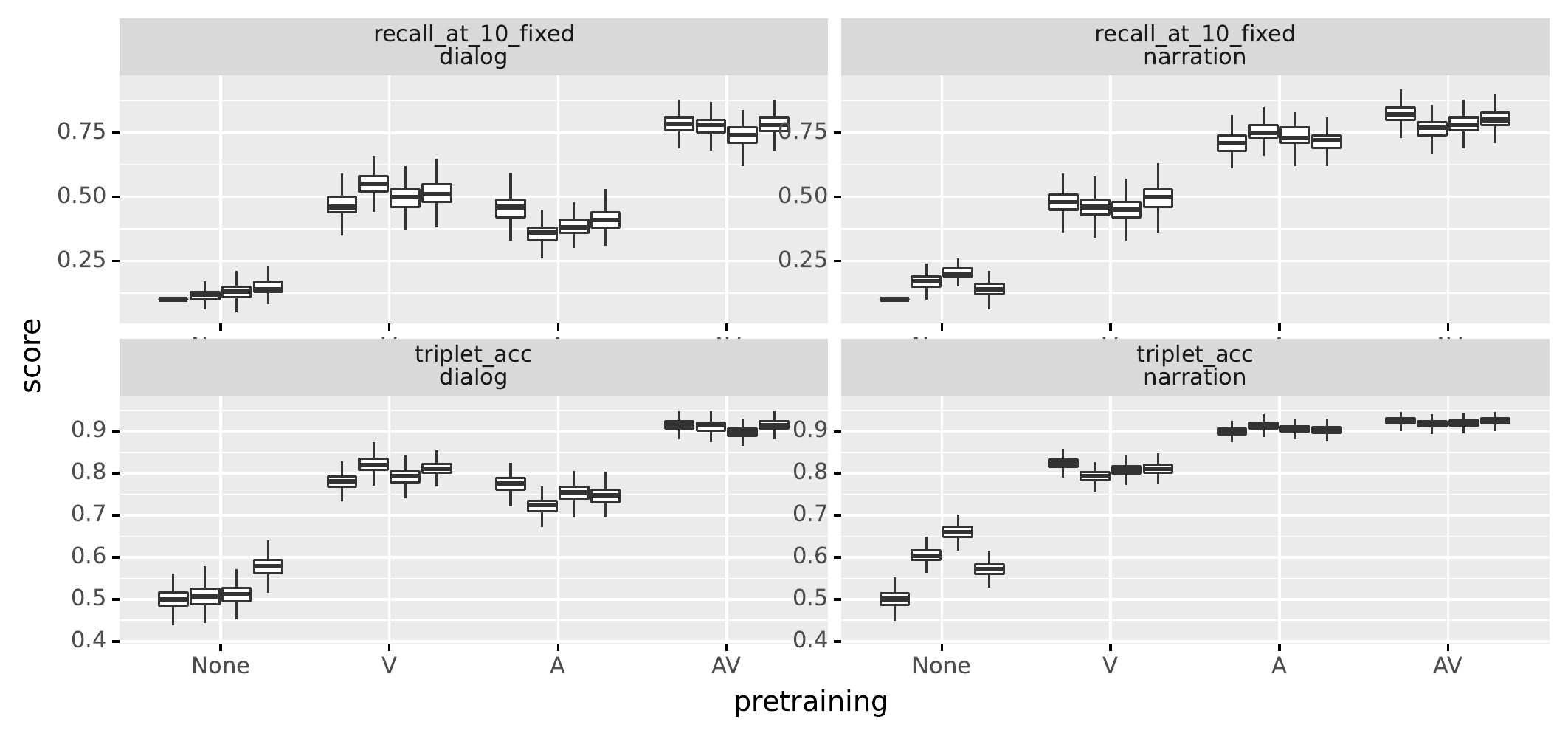}
	\caption{Effect of pretraining on performance on the dialog
          and narration validation data. The top row shows recall@10
          (chance = 10\%); the bottom row triplet accuracy (chance =
          50\%). Within each condition, we show scores for four
          separate runs. AV: pretrained audio and video; A: pretrained
          audio; V: pretrained video; None: no pretraining.}
	\label{fig:pretraining}
      \end{figure*}

Results on different pretraining configurations are shown in
\Cref{fig:pretraining}.
The best overall performance on both the dialog and the narration data is 
achieved with a model where both the video and audio encoder are pre-trained 
before being fine-tuned on our data. On narration data, for both metrics,
we see a clear ranking of
configurations from best to worst: audio and video pretraining (AV), 
audio pretraining (A), video pretraining (V) and no training (None). 
Meanwhile for dialog data, the performance between A and V conditions
is comparable. In the absence of any pretraining (None),
some runs fail to converge, thus performing at chance level.

\begin{figure}[htb]
  \centering
  \includegraphics[width=\columnwidth]{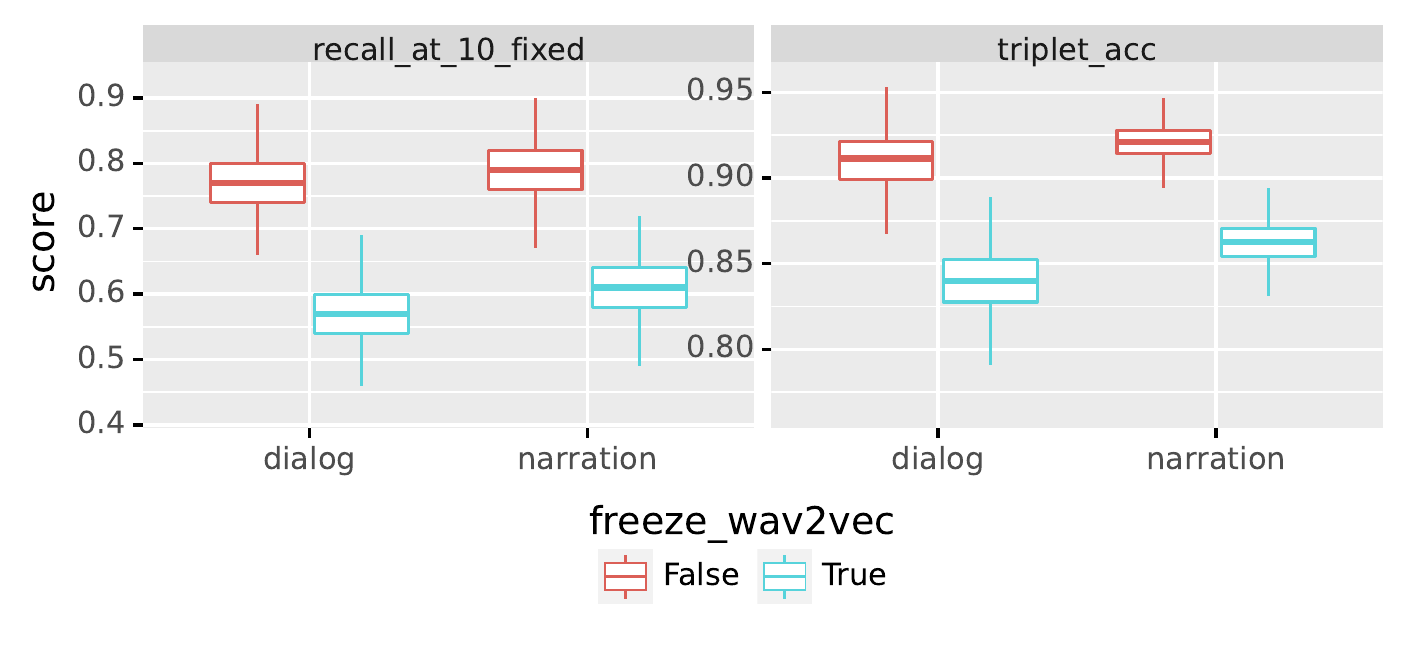}
  \caption{Effect of freezing the parameters of the \textsc{wav2vec2}
    module on model performance, on the dialog and narration
    validation data (True: \textsc{wav2vec2} frozen; False:
    \textsc{wav2vec2} trained). The left graph
    shows recall@10; the right graph triplet accuracy.}
  \label{fig:freeze_wav2vec}
\end{figure}

To further understand and disentangle the effects of audio pretraining and 
fine-tuning, we train a model with frozen parameters of the 
\textsc{wav2vec2} module. The effect of this condition is shown in \Cref{fig:freeze_wav2vec}.
We find without fine-tuning of the \textsc{wav2vec2} module, performance 
decreases substantially on both metrics. In other words, best 
performance is only achieved with pre-trained and fine-tuned models.

\subsubsection{Jitter}
\begin{figure}[!htb]
	\centering
	\includegraphics[width=\columnwidth]{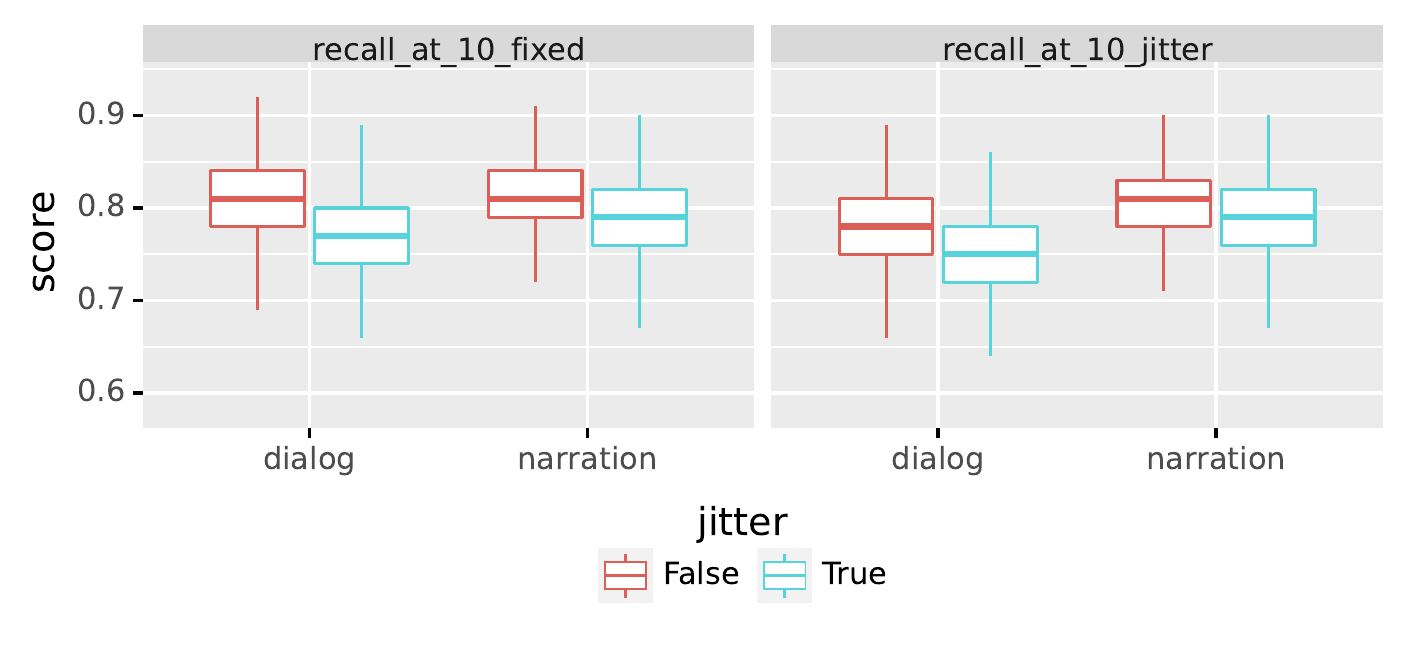}
	\caption{Effect of jitter on model performance, on the dialog
          and narration validation data (True: jitter; False:
          fixed). The left graph shows recall@10 on {\sc fixed}
          evaluation data; the right graph on {\sc jitter}-ed data.}
	\label{fig:jitter}
\end{figure}

Next, we evaluate a model that has been trained with varying video and audio 
lengths (\textsc{jitter}). For fair comparison, here we report recall@10 for both 
\textsc{fixed} and \textsc{jitter} validation configurations.
As seen in \Cref{fig:jitter}, the effect of \textsc{jitter} is only
minor and the performance is comparable.
However, we do observe some 
performance improvements when using \textsc{jitter} in the
minimal pairs evaluation (cf. \Cref{sec:minimal-pairs}).

\subsubsection{Temporal Information}
\begin{figure}[htb]
  \centering
  \includegraphics[width=\columnwidth]{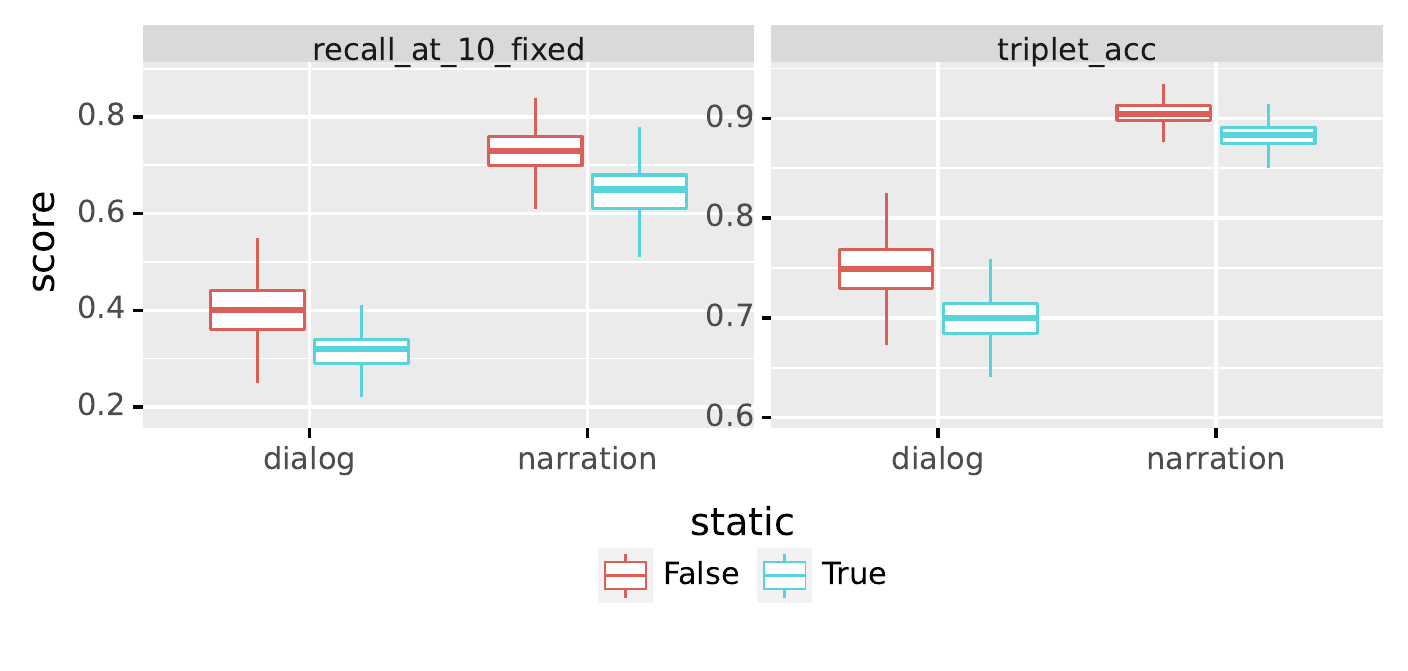}
  \caption{Effect of a \textsc{static} image encoder on model
    performance, on the dialog and narration validation data (True:
    static video encoder; False: regular video encoder). The left graph
    shows recall@10; the right graph triplet accuracy. For both conditions only
    the audio modality is pretrained.}
  \label{fig:static}
\end{figure}

\begin{figure}[htb]
  \centering
  \includegraphics[width=\columnwidth]{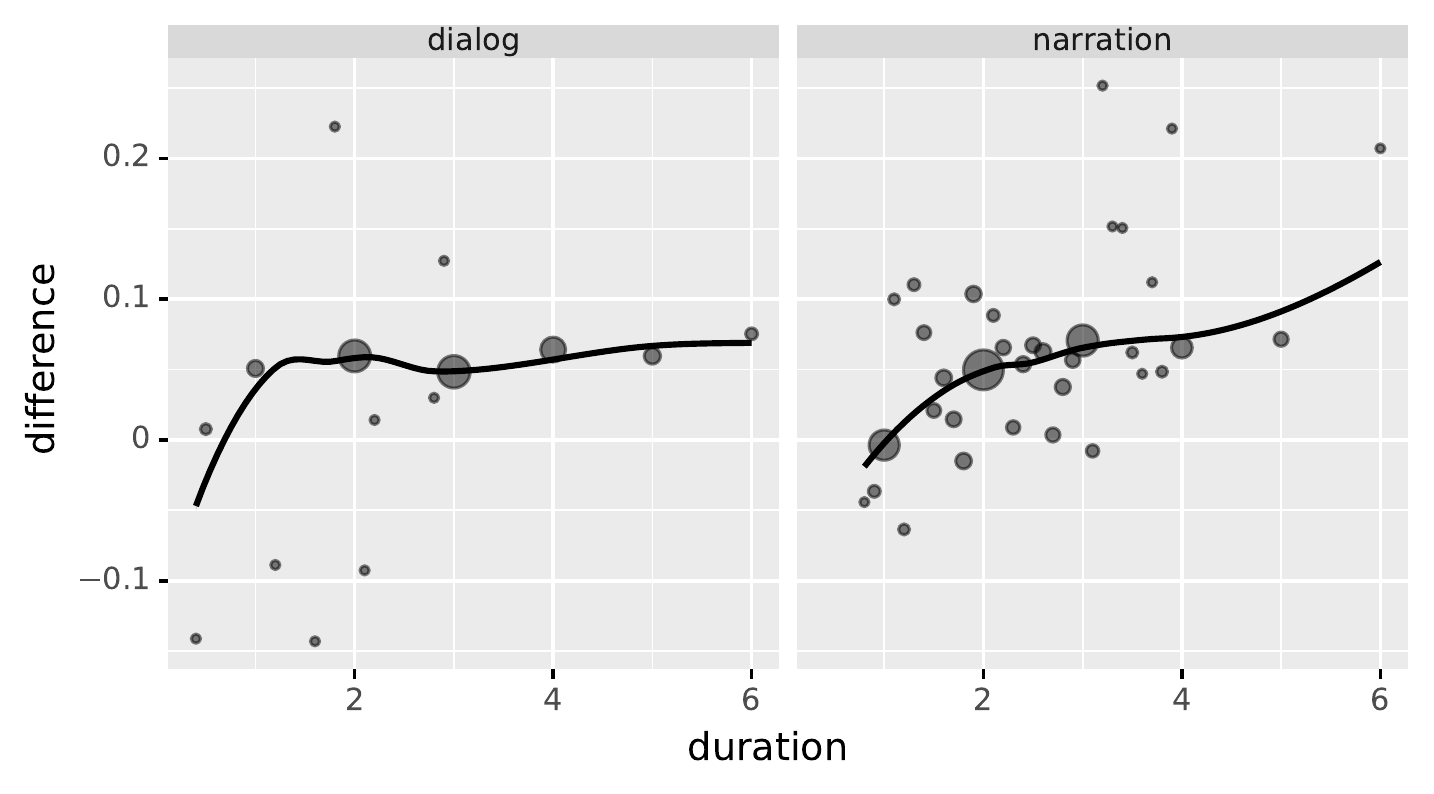}
  \caption{The effect of clip duration on the difference in
    mean score between models with/without access to temporal
    information, on triplet data. We calculate the 
    undiscretized triplet scores ($\mathrm{cosine}(A, P) - \mathrm{cosine}(A,
    N))$, average them over all same-duration triplets, and
    for each duration compute the difference in the average between
    time-aware and static models. Point size corresponds
    to the number of triplets within each duration. The line of fit is
    a {\sc loess} smoother weighted by size.}
  \label{fig:duration_effect}
\end{figure}
\begin{figure}[htb]
	\centering
	\includegraphics[width=\columnwidth]{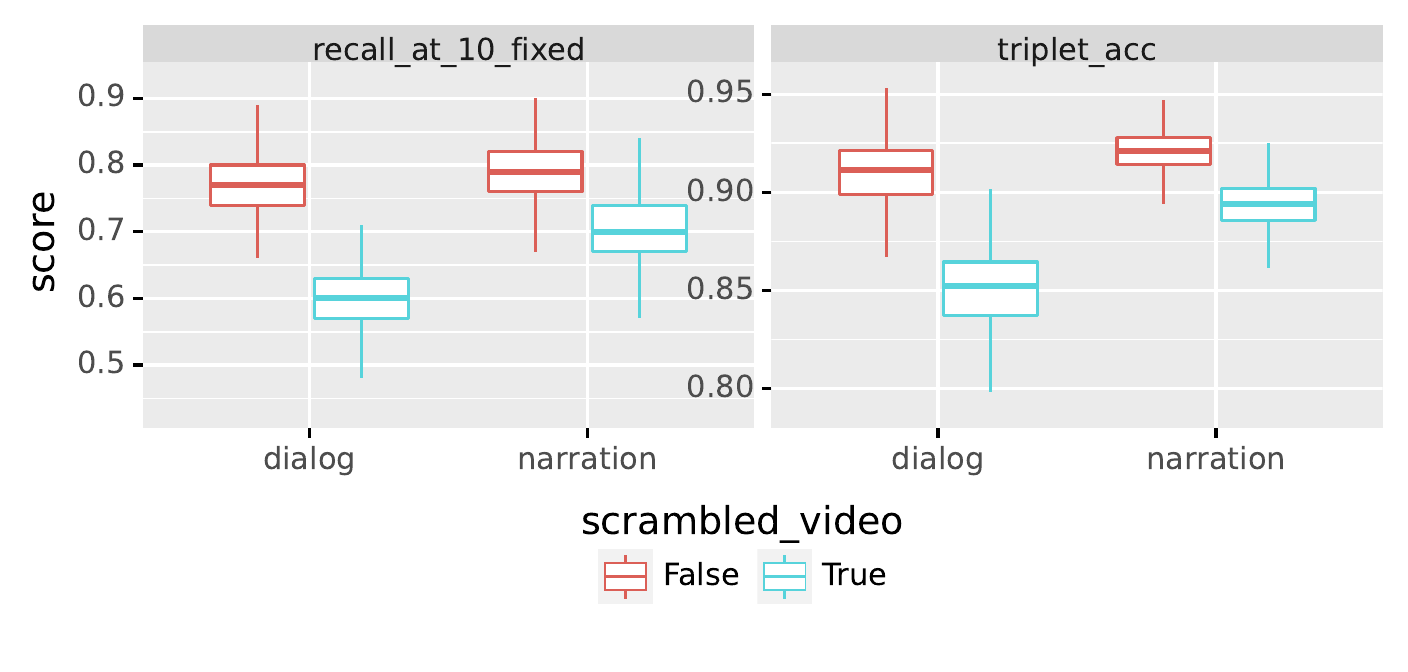}
	\caption{Effect of scrambling the video frames on model performance, on the 
	dialog and narration validation data (True: video frames scrambled;
        False: video frames in order). The left graph shows recall@10;
		the right graph triplet accuracy.}
	\label{fig:scrambled_video}
      \end{figure}
Finally, we explore the role of the temporal nature of the visual
modality.  \Cref{fig:static} compares the model with the regular video
encoder with one using the \textsc{static} baseline encoder.  For this
comparison we did not pretrain the video encoder in either condition,
in order to remove the confound of the pretraining data.\footnote{Note
  that there is one further confound we do not control for: the
  regular encoder has many more parameters than the {\sc static} one (31.5M vs. 11.7M).}
Across all
metrics, we observe substantial performance drops for the
\textsc{static} model, which has access to the same video frames, but
does not have access to their temporal ordering.

Additionally we investigate the effect of clip duration on this same
comparison, using the triplet evaluation
data. \Cref{fig:duration_effect} shows that the effect is nonlinear,
and for the shortest clips temporal information does not
help and may even have a detrimental effect.

\Cref{fig:scrambled_video} shows the effect of scrambling the video frames 
along the temporal dimension at test time (note that here the video encoders 
are pretrained). As expected, we observe substantial 
performance drops when the model does not see the video frames in 
the correct order. 

For this ablation the differential impact of clip duration on
the two conditions is very similar as in the {\sc static}
ablation (figure not included).

\subsection{Minimal Pairs}
\label{sec:minimal-pairs}
\begin{table*}[!htb]
	\centering
	\begin{tabular}{ccccccccc}
	\toprule
	ID & W2V Finet. &     Jitter &   V Pretr. &   A Pretr. &   Tmp Enc. & Tmp 
	Frames &     Nouns &     Verbs \\
	\midrule
	0 & \checkmark & \checkmark & \checkmark & \checkmark & \checkmark & 
	\checkmark & 0.80±0.02 & 0.79±0.02 \\
	1 &            & \checkmark & \checkmark & \checkmark & \checkmark & 
	\checkmark & 0.72±0.01 & 0.71±0.01 \\
	2 & \checkmark &            & \checkmark & \checkmark & \checkmark & 
	\checkmark & 0.72±0.02 & 0.78±0.01 \\
	3 & \checkmark & \checkmark &            &            & \checkmark & 
	\checkmark & 0.56±0.07 & 0.56±0.07 \\
	4 & \checkmark & \checkmark & \checkmark &            & \checkmark & 
	\checkmark & 0.69±0.02 & 0.69±0.01 \\
	5 & \checkmark & \checkmark &            & \checkmark & \checkmark & 
	\checkmark & 0.75±0.01 & 0.75±0.01 \\
	6 & \checkmark & \checkmark &            & \checkmark &            & 
	\checkmark & 0.78±0.01 & 0.76±0.01 \\
	7 & \checkmark & \checkmark & \checkmark & \checkmark & \checkmark 
	&            & 0.79±0.02 & 0.78±0.02 \\
	\bottomrule
\end{tabular}

	\caption{Minimal pair accuracies for nouns and verbs for different model 
		ablations. W2V Finet: \textsc{wav2vec2} module finetuned; A Pretr: 
		Audio encoder pretrained; V Pretr: Video encoder pretrained; Tmp 
		Enc: Video encoder with temporal information (not \textsc{static}); 
		Tmp Frames: Video frames in correct temporal order (not scrambled). 
		Mean and standard 
		deviation calculated over bootstrapped scores (100 re-samples), pooled 
		over 4 training runs.}
	\label{tab:minimal_pair_results}
\end{table*}
\begin{figure}[!htb]
  \centering
  \includegraphics[width=.5\textwidth]{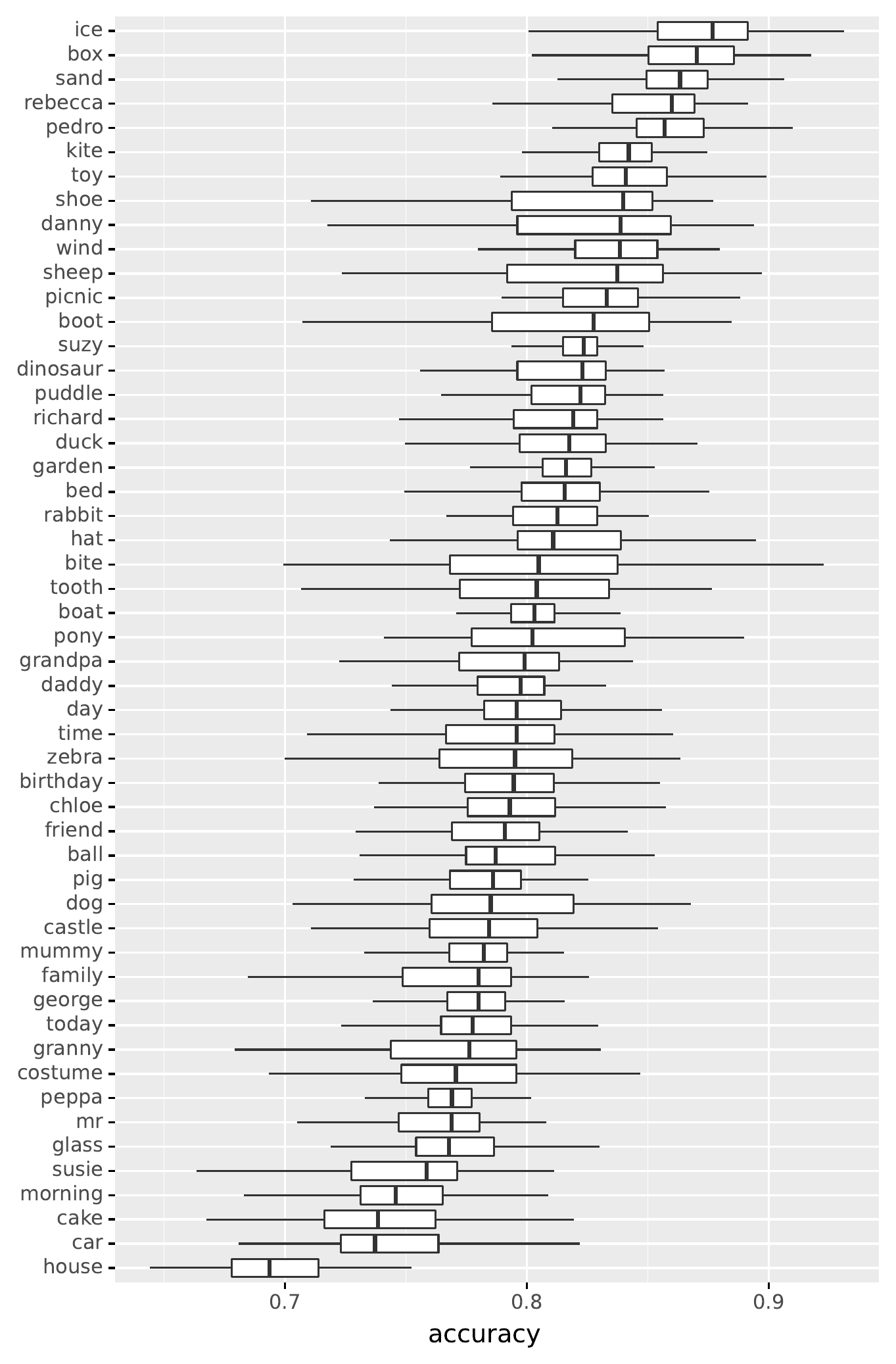}
  \caption{Per-word accuracies on the minimal pairs evaluation data for nouns.}
  \label{fig:accuracy_targeted_triplets_nouns}
\end{figure}
\begin{figure}[htb]
	\centering
	\includegraphics[width=.5\textwidth]{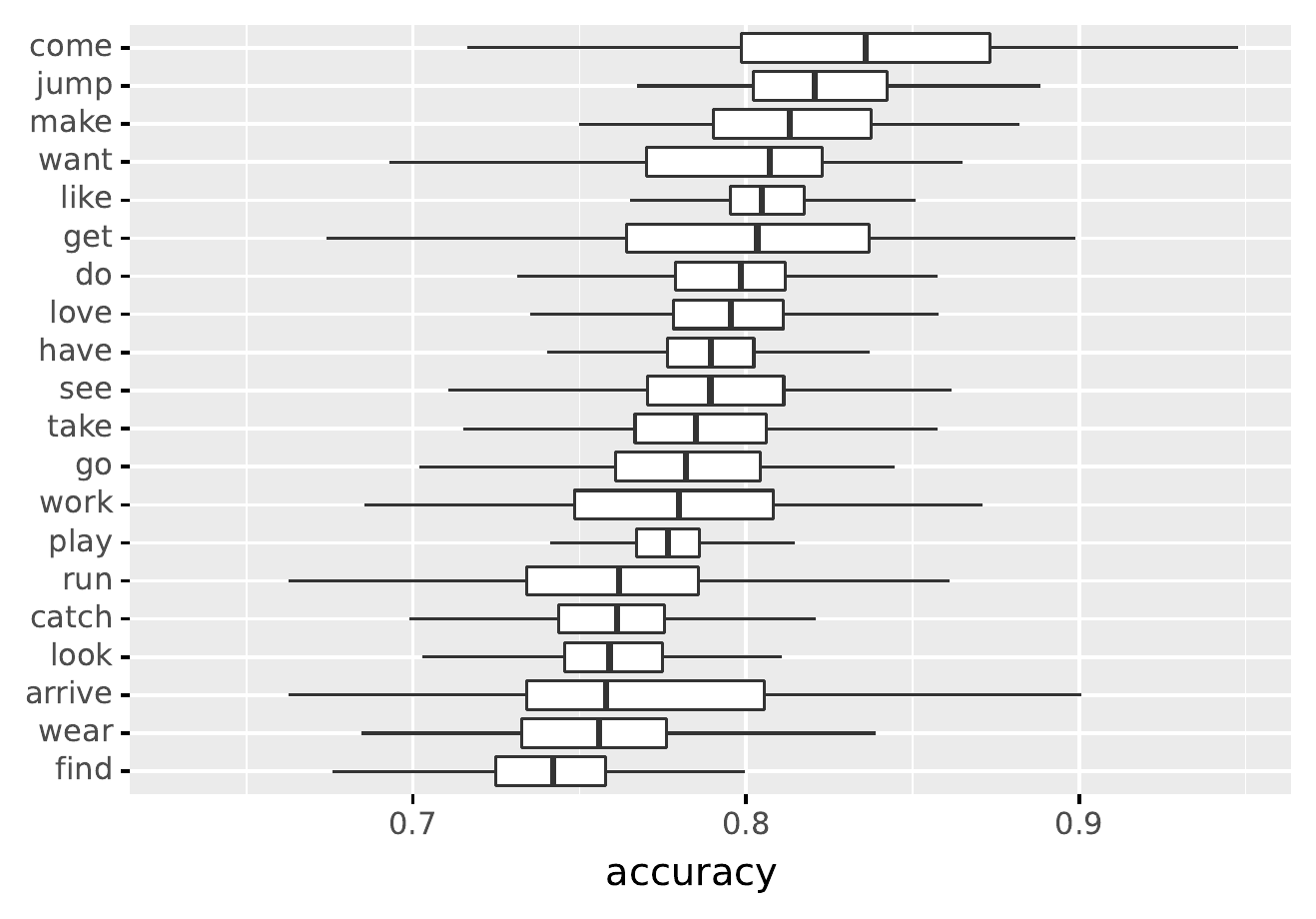}
	\caption{Per-word accuracies on the minimal pairs evaluation data for 
	verbs.}
	\label{fig:accuracy_targeted_triplets_verbs}
\end{figure}

\Cref{tab:minimal_pair_results} presents results for the minimal pair 
evaluation along with several ablations. Models which are 
pretrained and fine-tuned with \textsc{jitter} (row 0) perform best. In the 
first two configurations (rows 0 and 1), there is not much difference in the scores for 
verbs and nouns. However, we observe a substantial performance drop for both nouns and verbs if 
the \textsc{wav2vec2} module is not fine-tuned.

If the model is trained without \textsc{jitter} (row 2), performance drops substantially
for nouns, but not for verbs. One possible explanation for this could be that 
the evaluation samples for nouns are on average shorter than those for verbs 
(nouns: $0.43s$ vs. verbs: $0.49s$), and a model trained with \textsc{jitter} 
performs better on short clips because it has been exposed to clips of varying 
duration during training. Supporting this hypothesis, we find a positive 
correlation between log duration of clips and accuracy, which is lower for 
models trained with \textsc{jitter} (Pearson $r= 0.52$, $p < 0.001$) than for 
models without \textsc{jitter} (Pearson $r= 0.69$, $p < 0.001$).

In line with the general results, we find that the benefit of audio pretraining 
(row 5) is greater than that of video pretraining (row 4). A model without any 
pretraining (row 3) only performs marginally above chance.

For a model trained with a \textsc{static} video encoder (row 6), we compare 
performance to a model that was also trained without video pretraining (row 5) 
as done for the general results. We observe a slight performance 
improvement for nouns, and no significant difference for
verbs.
We suspect that temporal information is not crucial for the minimal pairs evaluation, 
because most evaluation samples are clips of short duration (on average: 
$0.44s$, i.e.\ 4-5 frames), thus limiting the benefit of the time dimension.
As we saw in the analysis of clip duration (\Cref{fig:duration_effect}), 
temporal information for such short clips does not improve performance, and 
could even have detrimental effects. In the alternative temporal ablation 
with scrambled video frames (row 7), we observe no significant performance 
drop compared to the base condition (row 0).

Figures \ref{fig:accuracy_targeted_triplets_nouns} and 
\ref{fig:accuracy_targeted_triplets_verbs} show per-word
accuracy for nouns and verbs for the best performing model configuration.
We observe substantial variance in the accuracy scores, suggesting that the 
difficulty to learn certain words varies. For example, the 
scores for \textit{house}, \textit{car}, and \textit{cake} are the lowest. This could be 
because these concepts are not easy to ground, either because they are used in 
displaced speech or because they do not often refer to a similar visual entity. 
When looking at our evaluation samples, we find that indeed the word \textit{house} 
is used in varying visual contexts (house entrance, whole house, inside the
house, rabbit's house) and in displaced speech (talking about going 
to somebody's house). Cars are only sometimes completely visible, often we see 
only cartoon characters \textit{in} a car. Regarding \textit{cake}, it refers to either a 
whole cake, a slice, dough, or crumbs.

On the other end, performance for concrete words such as \textit{ice}, 
\textit{box}, 
and \textit{sand} is the best, and indeed we find that in the evaluation examples 
these concepts are always present in the corresponding video and visually 
highly similar. Additionally, the words \textit{Pedro}, and \textit{Rebecca} 
are learned very well: They refer to \textit{Pedro Pony} and \textit{Rebecca Rabbit}, 
easily visually distinguishable from characters belonging to other species.

Further investigations with larger datasets are necessary to reveal the 
underlying reasons for difficulty, and relating them to predictors of age of 
acquisition in the child language acquisition literature 
\cite{roy2015predicting,frank2021variability}.

\section{Conclusion}
\label{sec:conclusion}
We simulate grounded language learning in a naturalistic setting, where 
the connection between the linguistic and visual modalities is not always strong 
and is potentially confounded by correlations with non-semantic aspects of 
the speech signal. Our experimental results suggest that despite the 
challenges inherent to the naturalistic aspects of our training dataset, a 
simple bimodal architecture can capture aspects of visual meaning of individual 
words as well as full utterances, and generalize well to narrative utterances
featuring a single unseen speaker and a descriptive rather than
conversational style. Our analyses show that generalization is substantially
boosted by fine-tuning audio representations pretrained on unlabeled
single-modality speech data. Fine-tuning a pretrained video encoder
also makes a contribution, but is less crucial to generalization from
dialog to narration.
We also investigate the role of temporal information in learning form-meaning 
mappings and show that having access to 
time information facilitates learning, except for very short video segments. 

\subsection{Limitations and Future Work}
To better understand what aspects of language are learning in our
setting, we need to to carry out in-depth analyses of learned 
representations on sub-word, lexical, and phrasal levels. It would also be 
worthwhile to figure out the details of how specifically temporal information 
in video contributes to acquiring linguistic knowledge.  Some analyses in 
this direction are currently constrained by the size of the evaluation 
dataset, and more large-scale datasets are needed in the future.

We model the acquisition of spoken language from 
language-internal correlations as well as from grounding in vision 
by fine-tuning an audio encoder pretrained on read speech. This 
approach is rather simplistic and does not match the real experience of 
language learners. It would be interesting to make the setting
more realistic by using pretraining data which reflect a young
learner's experience more closely, and to realistically interleave learning via
self-supervision from speech and via grounding in vision.
Ideally we would want to dispense with supervised pretraining of 
the video encoder as well and rather use a model pretrained in a
self-supervised way also for this modality.

\section*{Acknowledgements}
We would like to thank Nikos Papasarantopoulos and Shay B.\ Cohen for
creating the Peppa Pig annotations and for sharing them with us.
Part of the NWO/E-Science Center grant number 027.018.G03 was used to
purchase the video data on DVD. Thanks to Bertrand Higy for taking care of this
purchase, as well as for sharing his ideas with us in the initial stages of
this work. We also thank Abdellah Fourtassi, and three anonymous
reviewers for their useful feedback.

This work was supported by grants ANR-16-CONV-0002 (ILCB), AMX-19-IET-009 
(Archimedes Institute), and the Excellence Initiative of Aix-Marseille 
University (A*MIDEX).

\bibliography{biblio,anthology}
\bibliographystyle{acl_natbib}\balance
\end{document}